\documentclass[sigconf]{acmart}

\AtBeginDocument{%
  }

\copyrightyear{2025}
\acmYear{2025}
\setcopyright{acmlicensed}\acmConference[SIGIR '25]{Proceedings of the 48th International ACM SIGIR Conference on Research and Development in Information Retrieval}{July 13--18, 2025}{Padua, Italy}
\acmBooktitle{Proceedings of the 48th International ACM SIGIR Conference on Research and Development in Information Retrieval (SIGIR '25), July 13--18, 2025, Padua, Italy}
\acmDOI{10.1145/3726302.3731969}
\acmISBN{979-8-4007-1592-1/2025/07}

\usepackage{makecell}
\usepackage{booktabs}
\usepackage{siunitx}
\usepackage{balance}

\newcommand{\maxc}[1]{\textbf{#1}}

\begin{document}

\title{Data-efficient Meta-models for Evaluation of Context-based Questions and Answers in LLMs}

\author{Julia Belikova}
\orcid{0009-0007-7829-1249}
\affiliation{%
  \institution{Sber AI Lab}
  \city{Moscow}
  \country{Russia}
}
\affiliation{%
  \institution{Moscow Institute of Physics and Technology}
  \city{Dolgoprudny}
  \country{Russia}
}
\email{ju.belikova@gmail.com}

\author{Konstantin Polev}
\authornote{Corresponding author}
\orcid{0000-0002-0504-5940}
\affiliation{%
  \institution{Sber AI Lab}
  \city{Moscow}
  \country{Russia}
}
\email{endless.dipole@gmail.com}

\author{Rauf Parchiev}
\orcid{0009-0003-1678-1243}
\affiliation{%
  \institution{Sber AI Lab}
  \city{Moscow}
  \country{Russia}
}
\email{rauf.parchiev@gmail.com}

\author{Dmitry Simakov}
\orcid{0009-0003-3199-479X}
\affiliation{%
  \institution{Sber AI Lab}
  \city{Moscow}
  \country{Russia}
}
\email{dmitryevsimakov@gmail.com}

\renewcommand{\shortauthors}{Julia Belikova, Konstantin Polev, Rauf Parchiev, and Dmitry Simakov}

\begin{abstract}
Large Language Models (LLMs) and Retrieval-Augmented Generation (RAG) systems are increasingly deployed in industry applications, yet their reliability remains hampered by challenges in detecting hallucinations. While supervised state-of-the-art (SOTA) methods that leverage LLM hidden states—such as activation tracing and representation analysis—show promise, their dependence on extensively annotated datasets limits scalability in real-world applications. This paper addresses the critical bottleneck of data annotation by investigating the feasibility of reducing training data requirements for two SOTA hallucination detection frameworks: Lookback Lens, which analyzes attention head dynamics, and probing-based approaches, which decode internal model representations. We propose a methodology combining efficient classification algorithms with dimensionality reduction techniques to minimize sample size demands while maintaining competitive performance. Evaluations on standardized question-answering RAG benchmarks show that our approach achieves performance comparable to strong proprietary LLM-based baselines with only 250 training samples. These results highlight the potential of lightweight, data-efficient paradigms for industrial deployment, particularly in annotation-constrained scenarios. 

\end{abstract}

\begin{CCSXML}
<ccs2012>
   <concept>
       <concept_id>10010147.10010178.10010179.10010182</concept_id>
       <concept_desc>Computing methodologies~Natural language generation</concept_desc>
       <concept_significance>500</concept_significance>
       </concept>
 </ccs2012>
\end{CCSXML}

\ccsdesc[500]{Computing methodologies~Natural language generation}

\keywords{retrieval-augmented generation; question-answering; hallucination detection; data efficiency; model probing}

\maketitle

\section{Introduction}

Large Language Models (LLMs) and Retrieval-Augmented Generation (RAG) systems have rapidly become key components in diverse industry applications, including customer support automation and enterprise knowledge management. Despite their growing adoption, the reliability of these systems remains compromised by hallucinations: outputs that appear plausible but are factually incorrect or unsupported by the provided context. Such hallucinations pose significant risks in high-stakes domains such as healthcare, financial services, and legal advisory, undermining users' general trust.

Current state-of-the-art methods for hallucination detection exploit LLM hidden state representations through sophisticated techniques such as attention-based activation tracing \cite{chuang2024lookback} or internal representation probing \cite{sky2024androids}. However, these powerful approaches typically depend on large-scale annotated datasets or intensive computations on proprietary LLMs, limiting practical deployment.

In practice, industrial deployments face three critical constraints.
\textbf{Limited Annotated Data:} Specialized domain data frequently requires costly and time-consuming manual annotation, restricting the availability of labeled examples.
\textbf{Computational Efficiency:} Proprietary LLMs (e.g., GPT-4) often introduce prohibitive latency and sample processing costs, impeding real-time deployments.
\textbf{Privacy and Data Sovereignty:} Sensitive enterprise data often cannot be sent to external APIs, making the reliance on locally executable open-source models important.

Existing approaches are still struggling to collectively address these constraints, creating a considerable gap between academic benchmarks and feasible industrial solutions. To bridge this gap, our research investigates practical strategies for significantly improving data efficiency and achievable quality without sacrificing simplicity or computational efficiency.

Our key contributions specifically include:
\begin{enumerate}
    \item We present a framework that adapts the most effective probing techniques via multi-strategy feature extraction with dimensionality reduction and effective tabular classifiers, and provide a protocol for evaluating hallucination detection methods across the full data scarcity spectrum (50-1000 samples) on question-answering RAG datasets. Through comprehensive experiments, we quantitatively establish the critical points for preserving quality and provide remarkable insights into the applicability of the methods.
    
    \item We demonstrate the effectiveness of TabPFNv2 \cite{hollmann2025accurate} -- a tabular foundation model leveraging in-context learning -- to achieve state-of-the-art or competitive hallucination detection performance under limited data scenarios (see Table~\ref{tab_overall_results}). To our knowledge, we are among the first to rigorously explore the adaptability of advanced tabular foundation models (TabPFNv2) within hallucination detection tasks, highlighting valuable synergies between foundational tabular methods and NLP classification applications.

    \item We demonstrate that employing relatively small LLMs as extractor models for hallucination detection in some scenarios is comparable or even outperforms specialized evaluation frameworks based on proprietary models.
\end{enumerate}

\begin{table}[!ht]
    \centering
    \begin{footnotesize}
    \begin{tabular}{l|c|c|c|c|c}
    \hline
        \textbf{Classifier} & \textbf{EManual} & \textbf{ExpertQA} & \textbf{RAGTruth} & \textbf{Average} & \textbf{Rank} \\ \hline
        tabpfn              & \maxc{0.7161} & \underline{0.8204} & \underline{0.8139} & \maxc{0.7834} & 1 \\
        logreg              & \underline{0.6896} & \maxc{0.8218} & 0.8087        & \underline{0.7734} & 2 \\
        catboost            & 0.6832        & 0.7932        & \maxc{0.8176} & 0.7646        & 3 \\
        att.-pool. probe        & 0.6776        & 0.7611        & 0.8002        & 0.7463        & 4 \\ \hline
    \end{tabular}
    \end{footnotesize}
    \caption{Average ROC-AUC scores, corresponding classifier rankings, based on average on all datasets.}
    \label{tab_overall_results}
\end{table}

\vspace{-0.5cm}

The paper is structured as follows: Section~\ref{section:related} provides an overview of related work, Section~\ref{sec:framework} describes our proposed unified detection framework, Section~\ref{section:exp} presents experimental design and comprehensive experimental analysis, and Section~\ref{section:conclusion} concludes the paper with a summary of contributions and future research directions.

\section{Related Work}
\label{section:related}

\textbf{LLM hallucinations and their detection.} LLMs exhibit several types of hallucinations \cite{sahoo2024comprehensive, ji2023survey}. 
\textit{Factual hallucinations} \cite{zhang2023siren} involve generating information that conflicts with established facts. \textit{Semantic distortion} \cite{tjio2022adversarial} refers to errors where the meaning of a word or phrase is misrepresented. \textit{Contextual hallucinations} \cite{zhang2023siren} occur when the model's response is not consistent with a correctly provided context. In this work, the focus is specifically on contextual hallucinations in one of the most common industrial tasks -- question-answering (QA), where the question is augmented with grounding context. Detecting such hallucinations is a crucial step toward hallucination reduction in industrial systems, especially for applications requiring exact factual answers, such as medical \cite{frisoni2022bioreader, naik2022literature} or financial \cite{zhao2024optimizing, sarmah2024hybridrag, chen2024knowledge} RAG systems. 

This focus on contextual hallucination detection has led to the development of strong context-based methods such as Lookback Lens, introduced in \cite{chuang2024lookback}, which measures the \textit{lookback ratio} of attention weights between the provided context and newly generated tokens, applying a linear classifier to these features. Other work \cite{orgad2024llmsknowshowintrinsic, sky2024androids} shows the high effectiveness of different probing methods for domain-specific learning, such as \cite{sky2024androids}, which trains \textit{linear}, \textit{attention-pooling}, and \textit{ensemble probes} over the outputs of a transformer block~\cite{vaswani2017attention}. A parallel line of research explores uncertainty estimation for hallucination detection \cite{fomicheva2020unsupervised, kuhn2023semantic, manakul2023selfcheckgpt, duan-etal-2024-shifting}. These methods are rapidly evolving but still may struggle with calibration issues, require multiple forward passes, or may not capture the nuanced relationship between model confidence and factual accuracy \cite{lookbeforeyouleap}. Therefore, in this work, we focus on lightweight probing techniques that have proven to be effective in industrial domains~\cite{baek-etal-2025-probing}.

The recent advancements in LLMs \cite{achiam2023gpt, liu2024deepseek} have also enabled researchers to use LLMs as judges to evaluate other models' answers \cite{zheng2023judging}. Notably, RAGAS \cite{es2024ragas}, a comprehensive evaluation system that uses an LLM evaluator, is specifically designed for RAG systems. While these and other frameworks like \cite{saad2024ares, belyi2025luna} and TruLens\footnote{\url{https://www.trulens.org}} have introduced powerful methods for the contextual hallucination detection task, they typically require access to multiple model responses, large amounts of data, or access to proprietary models. In contrast, our methodology is designed for data- and compute-constrained settings, making it more suitable for industrial deployment.

\textbf{In-context learning for tabular data.} Recent advances in deep learning have extended to tabular tasks, traditionally solved using linear or gradient boosting algorithms \cite{prokhorenkova2018catboost}. In-context learning approaches are particularly effective in low-data scenarios, such as when limited annotated data is available for hallucination detection. TabPFN \cite{hollmann2022tabpfn, hollmann2025accurate}, a tabular foundation model designed for small datasets, has shown promise in this area, outperforming previous methods while requiring significantly less computation time. This paper investigates TabPFN's applicability to hallucination detection and compares its performance with other tabular classifiers.

\section{Framework}
\label{sec:framework}

Our framework introduces a unified and data-efficient approach to contextual hallucination detection in question-answering tasks. This two-stage pipeline combines feature extraction from LLM internals with lightweight classification to maximize detection performance while minimizing training data requirements.

\textbf{Feature extraction}. An \textit{annotator} LLM generates an answer to a contextual augmented question, while an independent \textit{extractor} LLM computes internal activations -- specifically hidden states (outputs of transformer blocks) and attention scores. This separation is necessary when the hidden states of the generator model are inaccessible. Further, we employ two complementary approaches to aggregate and compress the extracted activations:
\begin{enumerate}
    \item For each sequence of per-token hidden states, we apply mean and max pooling across the feature dimension and extract the last token's hidden state, which research has shown to be particularly informative for hallucination detection~\cite{sky2024androids}. These three components are taken from the middle layer hidden states (based on empirical findings from~\cite{chuang2024lookback}) and compressed separately using dimensionality reduction techniques (PCA or UMAP) to a fixed size, a constraint motivated by our goal of minimal training data requirements. The resulting vectors are then concatenated to form a comprehensive feature representation.

    \item Building on the Lookback Lens methodology, we compute the ratio of attention weights between the provided context and newly generated tokens. Although the original implementation applies this analysis to specific token spans, we adapt it by computing the mean lookback ratio across the entire sequence, making it more suitable for our data-efficient paradigm. Given the high dimensionality of lookback ratios (layers × attention heads), we propose two alternative feature extraction strategies: applying dimensionality reduction similar to the pooling probe approach and selectively using the middle layers of the model with all their attention heads while ensuring the feature count remains within the input constraints of our tabular classifiers (particularly the TabPFNv2's 500-feature limit). 
\end{enumerate}

\textbf{Meta-Classification}.
The extracted feature vectors are fed into lightweight meta-classifiers specifically chosen for their effectiveness in low-data settings. We evaluated three primary classification approaches: logistic regression, CatBoost~\cite{prokhorenkova2018catboost}, and TabPFN2.

\section{Experiments}
\label{section:exp}

\subsection{Experimental Setup}
\textbf{Datasets}. Three datasets that reflect real-world industrial constraints are used in this study: RAGTruth~\cite{niu2023ragtruth} (QA subset), and ExpertQA~\cite{malaviya2023expertqa} and EManual~\cite{emanual} datasets from RAGBench~\cite{friel2024ragbench}. RAGTruth is a human-annotated dataset with labels curated by domain experts, whereas RAGBench uses GPT-4 as the annotator to generate high-quality, scalable, albeit synthetic, hallucination labels. Examples of context-based question-answer pairs from these datasets are shown in Table~\ref{tab:dataset-example}.

\begin{table}[H]
\small
\centering
% \rowcolors{1}{gray!15}
\begin{tabular}{p{0.95\linewidth}}
    \toprule
    \textbf{RAGTruth}: \color{red}{\textit{hallucinated answer}}\\
    \midrule
    \textbf{Question}: Risks benefits roth ira vs 401k \\
    \textbf{Context}: Passage [1]: A Roth 401(k) account works much like a Roth IRA, but there is no income limit to participate, and you are required to take the minimum distributions ... \\
    \textbf{Answer}: Based on the given passages, the benefits of a Roth IRA include the lack of income limits to participate, the ability to save more after maxing out the Roth IRA contribution, and the higher contribution limits compared to a Roth IRA ... \\
    \midrule 
    
    \textbf{ExpertQA}: \color{teal}{\textit{correct answer}}\\
    \midrule
    \textbf{Question}: What is the best way to store my medication to ensure its potency and safety? \\
    \textbf{Context}: Document [1]: Maintaining the right temperature in a pharmacy refrigerator is crucial to ensure the safe storage of medication ... \\
    \textbf{Answer}: The best way to store medication to ensure its potency and safety is to follow the storage instructions provided ... \\
    \midrule 

\end{tabular}
\caption{Examples of context-based question-answer pairs from RAGTruth and ExpertQA datasets.}
\vspace{-10pt}
\label{tab:dataset-example}
\end{table}

\textbf{Training and Evaluation}. The performance of methods is evaluated under the available training data size constraint. For RAGBench, training subsets of sizes 50, 100, 250, 500, 750, and 1000 are sampled, while for RAGTruth, we explore sizes 50, 100, 250, and 500. In each case, we hold out 20\% of the training samples for validation and use the default test splits. Each experiment is repeated with different random seeds three times to ensure robust performance estimates. 

We use ROC-AUC as our primary evaluation metric, as it is independent of threshold and provides a robust measure of detection performance across different operating points. In addition, we report mean reciprocal rank (MRR) to evaluate how well our models prioritize the detection of hallucinations when ranking multiple responses (see Table~\ref{tab:3}).
% \vspace{-0.5cm}

\begin{table}[!ht]
    \centering
    \begin{tabular}{l|c|c|c}
    \hline
    \textbf{Classifier} &
      \makecell{\textbf{gemma-2-}\\\textbf{9b-it}} &
      \makecell{\textbf{Llama-3.1-}\\\textbf{8B-Instruct}} &
      \makecell{\textbf{Qwen2.5-}\\\textbf{7B-Instruct}}         \\ \hline
    att.-pool. probe & 0.2778        & 0.3611        & 0.3333        \\
    catboost     & 0.4444        & 0.3889        & \underline{0.5556} \\
    logreg       & \maxc{0.7778} & \underline{0.5}    & 0.4167        \\
    tabpfn       & \underline{0.5833} & \maxc{0.8333} & \maxc{0.7778} \\ \hline
    \end{tabular}
    \caption{Mean reciprocal rank across datasets for the detectors. Higher values indicate superior performance.}
    \label{tab:3}
\end{table}
\vspace{-0.5cm}

% \subsection{Methods Configuration}
\textbf{Baselines}. For comparison, the standard attention-pooling probe is evaluated against the proposed modifications. In addition, we evaluate the GPT-4o zero-shot judge and specialized prompting of the RAGAS framework with the GPT-4o model for the faithfulness assessment. As explained above, these approaches do not satisfy the constraints of the industry-related setting we consider but serve as practical upper bounds in our experiments.

\textbf{Main methods}. For the methods in the proposed framework, we use three smaller open-source LLMs as extractors of internal activations: Gemma-2-9B-It \cite{gemma_2024}, Llama-3.1-8B \cite{grattafiori2024llama}, and Qwen2.5-7B-Instruct \cite{qwen2.5}. 
When the dimensionality reduction step is applied, each feature type -- namely pooled hidden states and lookback ratios -- is reduced to 30 components to accommodate training size limitations. This ensures the total feature count remains below the 500 limit required for TabPFNv2 compatibility. Alternatively, when dimensionality reduction is omitted for lookback ratios, specific layer ranges (Qwen: 5–21, Llama: 8–22, Gemma: 5–35) are selected to approximate this 500-feature ceiling, while accounting for variations in attention head counts.

\subsection{Results and Analysis}

\begin{figure}[H]
    \centering
    \includegraphics[width=0.45\textwidth]{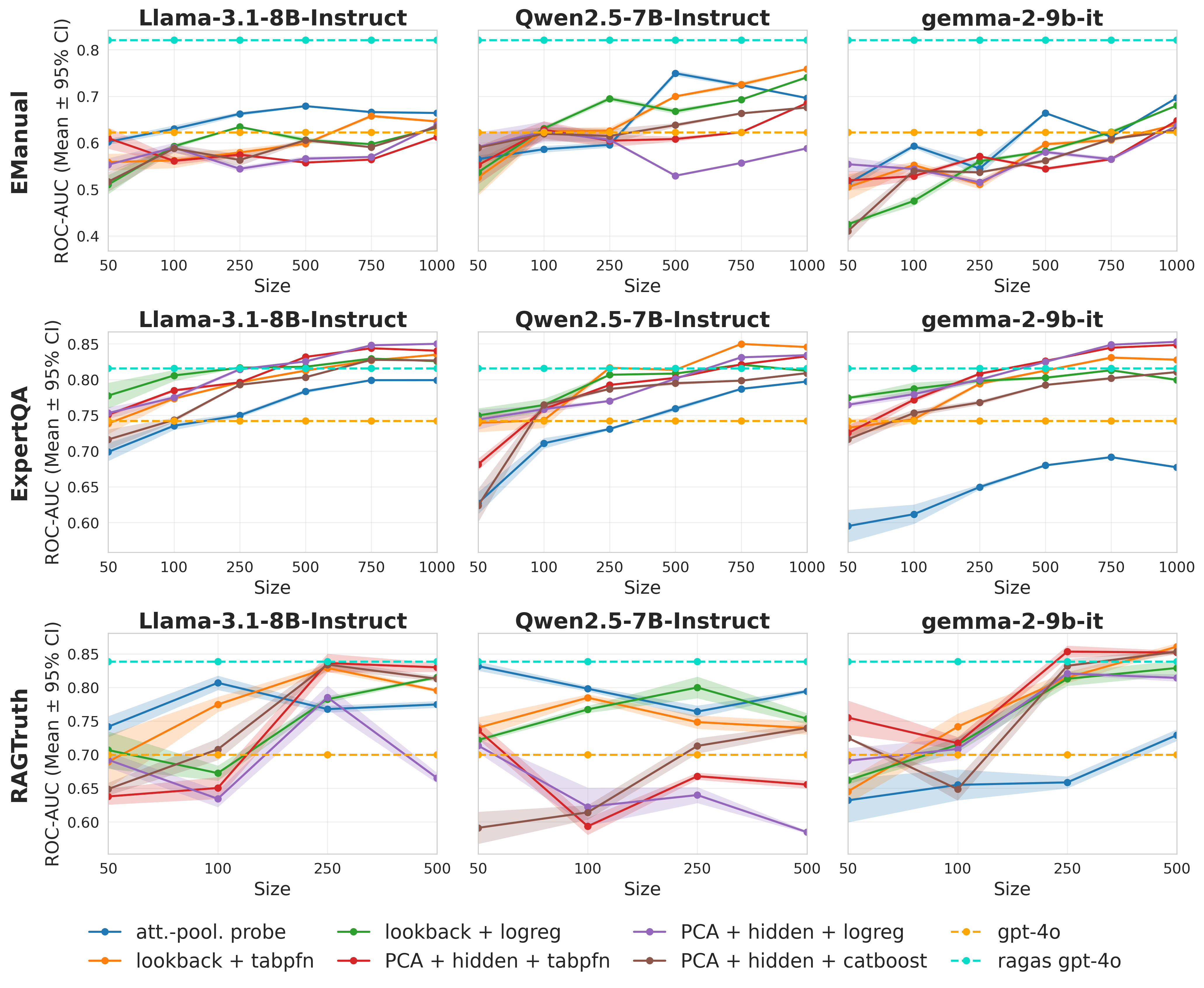}
    \caption{Test ROC‑AUC versus training‑set size for the proposed evaluators (solid lines) across the three benchmarks (rows) and three response generators (columns). Horizontal dashed lines correspond to the zero‑shot GPT‑4o judge (yellow) and the RAGAS GPT‑4o pipeline (cyan). Shaded areas indicate ±95\% confidence intervals over three random seeds.}
    \label{fig:curves}
    \Description{Test ROC‑AUC versus training‑set size plot.}
\end{figure}

The analysis explores findings across five directions: (1)~overall effectiveness concerning strong external baselines, (2)~impact of feature design and dimensionality reduction, (3)~choice of meta-classifier, (4)~choice of extractor LLM, and (5)~data-efficiency.  All numbers are averaged over three random seeds and are shown in Figure~\ref{fig:curves} for the main set of methods. Table~\ref{tab:2} gives a brief cross-extractor snapshot of detector quality before we examine the results in detail.

\begin{table}[!ht]
    \centering
    \begin{tabular}{l|c|c|c}
    \hline
        \textbf{Classifier} &       
        \makecell{\textbf{gemma-2-}\\\textbf{9b-it}} &
        \makecell{\textbf{Llama-3.1-}\\\textbf{8B-Instruct}} &
        \makecell{\textbf{Qwen2.5-}\\\textbf{7B-Instruct}} \\ \hline
        att.-pool. probe & 0.6379                           & 0.7226            & 0.7199             \\
        catboost     & 0.719                            & 0.7328            & 0.7466             \\
        logreg       & \maxc{0.7334}                    & \underline{0.7378}     & \underline{0.7505}      \\
        tabpfn       & \underline{0.7242}                    & \maxc{0.7587}     & \maxc{0.7623}      \\ \hline
    \end{tabular}
    \caption{Mean ROC-AUC across datasets for the detectors.}
    \label{tab:2}
\end{table}

\vspace{-0.2cm}

\begin{table}[!ht]
  \centering
  \sisetup{
    round-mode          = places,
    round-precision     = 4,
    table-number-alignment = center
  }
  \small
  \begin{tabular}{
      l
      S[table-format=1.4]
      S[table-format=1.4]
      S[table-format=1.4]
    }
    \toprule
    \textbf{Method}                & \textbf{EManual}    & \textbf{ExpertQA}   & \textbf{RAGTruth}  \\
    \midrule
    PCA + lookback + tabpfn        & \maxc{0.6972} & \maxc{0.8165} & 0.7679        \\
    lookback + tabpfn              & 0.6659        & 0.8064        & \maxc{0.8037} \\
    lookback + logreg              & 0.6608        & 0.8122        & 0.7828        \\
    att.-pool.\ probe                  & 0.6776        & 0.7611        & 0.8002        \\
    PCA + hidden + tabpfn          & 0.6266        & 0.8115        & 0.7944        \\
    PCA + lookback + catboost      & 0.6736        & 0.7785        & 0.7717        \\
    PCA + lookback + logreg        & \underline{0.6872} & 0.8051        & 0.7292        \\
    PCA + hidden + catboost        & 0.6336        & 0.7887        & 0.7801        \\
    PCA + hidden + logreg          & 0.6018        & \underline{0.8143} & 0.7643        \\
    lookback + catboost            & 0.5898        & 0.7792        & \underline{0.8028} \\
    UMAP + hidden + tabpfn         & 0.6714        & 0.7758        & 0.7131        \\
    UMAP + lookback + logreg       & 0.5960        & 0.7577        & 0.7701        \\
    UMAP + hidden + catboost       & 0.6479        & 0.7672        & 0.6987        \\
    UMAP + hidden + logreg         & 0.6439        & 0.7723        & 0.6899        \\
    UMAP + lookback + tabpfn       & 0.5941        & 0.7529        & 0.7462        \\
    UMAP + lookback + catboost     & 0.6226        & 0.7324        & 0.6941        \\ 
    \midrule
    RAGAS GPT-4o                   & 0.8208        & 0.8160        & 0.8386        \\
    GPT-4o                         & 0.6227        & 0.7423        & 0.7000        \\
    \bottomrule
  \end{tabular}
  \caption{Mean ROC-AUC scores across extractor models of various combinations of feature extraction,
    dimensionality reduction, and classification algorithms, along with the scores from two types of LLM judges.}
  \label{tab:1}
\end{table}

\vspace{-0.4cm}

\textbf{Overall effectiveness.}
Our experiments demonstrate that lightweight methods leveraging internal states of LLMs can achieve substantial hallucination detection performance, even with limited training data. As aggregated in Table~\ref{tab:1}, the best-performing methods, especially those combining lookback features with TabPFN, approach (for RAGTruth) or, in some cases, even surpass (ExpertQA) the performance of the strong RAGAS GPT-4o baseline. While RAGAS GPT-4o generally sets a high benchmark (ROC-AUC 0.81-0.84), the proposed methods reach competitive levels (up to 0.81 ROC-AUC). They significantly outperform zero-shot GPT-4o, validating their potential for efficient local hallucination detection.

\textbf{Impact of feature design and dimensionality reduction.} PCA dimensionality reduction often enhances lookback feature performance, proving more effective than UMAP or using raw features in most configurations.

\textbf{Choice of meta-classifier.} TabPFN is the top-performing meta-classifier on average ROC-AUC and MRR, demonstrating effectiveness in low-data settings (Tables~\ref{tab_overall_results}, \ref{tab:3}, \ref{tab:2}). Logistic Regression is also highly competitive. Interestingly, both, on average, outperform CatBoost and the baseline attention-pooling probe approach.

\textbf{Choice of extractor LLM.} We observe that the choice of extractor LLM does influence the results; Llama-3.1-8B and Qwen2.5-7B, on average, yield superior detector performance compared to Gemma-2-9B across classifiers (Figure~\ref{fig:curves}).

\textbf{Data-efficiency.} The methods exhibit strong data-efficiency (Figure~\ref{fig:curves}, Table~\ref{tab2}). Performance improves sharply with initial data (50-250 samples) before plateauing. High performance relative to baselines is achievable with only 250 examples, confirming suitability for annotation-constrained industrial settings.

\begin{table}[H]
    \centering
    \begin{tabular}{c|c|c|c}
    \hline
        \textbf{Train size} & \textbf{EManual} & \textbf{ExpertQA} & \textbf{RAGTruth} \\ \hline
        50 & 0.5246 & 0.6939 & 0.6604 \\
        100 & 0.5569 & 0.7294 & 0.6565 \\
        250 & 0.5903 & 0.7648 & 0.7368 \\
        500 & 0.6051 & 0.7809 & 0.7655 \\
        750 & 0.6056 & 0.7959 & - \\
        1000 & 0.6427 & 0.7985 & - \\ \hline
    \end{tabular}
    \caption{Effect of dataset size on quality (ROC-AUC).}
    \label{tab2}
\end{table}
\vspace{-0.5cm}

\textbf{In general}, we conducted a thorough evaluation involving multiple combinations of dimensionality reduction techniques, feature extraction methods, and meta-classifiers.  Our experiments indicate that methods based on TabPFNv2 exhibit strong performance in data-scarce settings and, on average, outperform other classifiers we evaluated.

\section{Conclusion}
\label{section:conclusion}
The reliability of LLM-driven RAG systems in industry settings hinges on scalable, cost-effective methods to detect hallucinations. This work demonstrates that lightweight, annotation-efficient approaches can achieve competitive performance while overcoming three critical industrial constraints: limited labeled data, prohibitive computational costs, and data privacy risks. Proposed data-efficient methods also highlight novel use cases for TabPFNv2 in NLP domains, reveal limitations of attention-based probing techniques, and illustrate effective, lightweight heuristic strategies for hallucination detection. Future research should further explore architectural modifications of tabular foundation models (e.g., TabPFNv2) specifically optimized for NLP representation learning and contextual hallucination detection tasks.

\begin{acks}
We thank Kseniia Kuvshinova and Aziz Temirkhanov for valuable discussions, feedback and their high contribution to this work.
\end{acks}
\newpage\section*{Presenter Bio}

Julia Belikova is an M.Sc. candidate in computer science at the Moscow Institute of Physics and Technology. She is also an NLP Researcher at Sber AI Laboratory. Her work focuses on addressing critical challenges in generative AI, particularly LLM hallucination detection and uncertainty quantification.

\bibliographystyle{ACM-Reference-Format}
\balance
\bibliography{references}

\end{document}